%% file: main.tex
\documentclass{article}

\usepackage{microtype}
\usepackage{graphicx}
\usepackage{subfigure}
\usepackage{booktabs} %

\usepackage{hyperref}

\usepackage{terms}

\usepackage[accepted]{icml2023}

\usepackage{amsmath}
\usepackage{amssymb}
\usepackage{mathtools}
\usepackage{amsthm}

\usepackage[capitalize,noabbrev]{cleveref}

\theoremstyle{plain}

\theoremstyle{definition}

\theoremstyle{remark}

\usepackage[textsize=tiny]{todonotes}

\icmltitlerunning{A Critical View of Vision-Based Long-Term Dynamics Prediction Under Environment Misalignment}

\begin{document}

\twocolumn[
\icmltitle{A Critical View of Vision-Based Long-Term Dynamics Prediction \\
Under Environment Misalignment}

\begin{icmlauthorlist}
\icmlauthor{Hanchen Xie}{isi,usccs}
\icmlauthor{Jiageng Zhu}{isi,uscee}
\icmlauthor{Mahyar Khayatkhoei}{isi}
\icmlauthor{Jiazhi Li}{isi,uscee}
\icmlauthor{Mohamed E. Hussein}{isi,alex}
\icmlauthor{Wael AbdAlmageed}{isi,usccs,uscee}
\end{icmlauthorlist}

\icmlaffiliation{isi}{USC Information Sciences Institute, Marina del Rey, USA}
\icmlaffiliation{usccs}{USC Thomas Lord Department of Computer Science, Los Angeles USA}
\icmlaffiliation{uscee}{USC Ming Hsieh Department of Electrical and Computer Engineering, Los Angeles, USA}
\icmlaffiliation{alex}{Alexandria University, Alexandria, Egypt}

\icmlcorrespondingauthor{Hanchen Xie}{hanchenx@isi.edu}

\icmlkeywords{Machine Learning, ICML}

\vskip 0.3in
]

\printAffiliationsAndNotice{}  %

\input{sections/00_abstract}

\input{sections/01_introduction}
\input{sections/02_related_works}
\input{sections/03_method}
\input{sections/04_experiments}

\input{sections/06_conclusion}
\input{sections/08_acknowledgements}

\bibliography{reference}
\bibliographystyle{icml2023}

\newpage
\appendix
\onecolumn
\input{sections/07_Appendix.tex}

\end{document}

%% file: sections/00_abstract.tex
\begin{abstract}

Dynamics prediction, which is the problem of predicting future states of scene objects based on current and prior states, is drawing increasing attention as an instance of learning physics. To solve this problem, Region Proposal Convolutional Interaction Network (RPCIN), a vision-based model, was proposed and achieved state-of-the-art performance in long-term prediction. RPCIN only takes raw images and simple object descriptions, such as the bounding box and segmentation mask of each object, as input. However, despite its success, the model's capability can be compromised under conditions of environment misalignment. In this paper, we investigate two challenging conditions for environment misalignment: \emph{Cross-Domain} and \emph{Cross-Context} by proposing four datasets that are designed for these challenges: \emph{SimB-Border}, \emph{SimB-Split}, \emph{BlenB-Border}, and \emph{BlenB-Split}. The datasets cover two domains and two contexts. Using RPCIN as a probe, experiments conducted on the combinations of the proposed datasets reveal potential weaknesses of the vision-based long-term dynamics prediction model. Furthermore, we propose a promising direction to mitigate the \emph{Cross-Domain} challenge and provide concrete evidence supporting such a direction, which provides dramatic alleviation of the challenge on the proposed datasets. 

\end{abstract}

%% file: sections/01_introduction.tex
\section{Introduction}
In addition to identifying the visual patterns observed in a scene, such as in object detection~\cite{fasterrcnn, yolo} and segmentation~\cite{fcn, maskrcnn}, recently, increasing attention has been drawn to learning the underlining mechanisms of scene generation, such as learning the laws of physics \cite{interaction_learn_phys, compo_obj_base_phys, clevrer,dynamic_reason_diff_phys}. An intuitive instance of learning physics is \textit{dynamics prediction}~\cite{compo_video_pred, rpcin}, in which the future states of discrete objects are predicted from the observations of reference states. A number of approaches has been proposed to solve the dynamics prediction problem. Object-centric models~\cite{compo_obj_base_phys, reasoing_phys_interaction} along with interaction networks~\cite{interaction_learn_phys, visual_interaction} focus on extracting representations and modeling dynamics of each object. 

\input{sections/figures_text/concept_figure.tex}

One stream of approaches models the dynamics by incorporating human defined physics models~\cite{mujuco, ullman2014learning,ete_diff_phy, dynamic_reason_diff_phys}, where the physics parameters of the reference states, such as position, mass, and velocity, can be extracted from visual information or given as prior knowledge. Despite their accuracy with the guidance of Newtonian physics theory, developing a physics model that precisely describes the mechanism underlying complex real-world scenario requires expert knowledge and sophisticated physics parameters which can be hard to acquire, such as energy lost during inelastic collision.
In an alternative stream of approaches, several works~\cite{galileo, physics_play_billiard, interaction_learn_phys, compo_obj_base_phys, visual_de_animation, compo_video_pred} rely on Deep Neural Network (DNN) for learning physics, where the underline mechanisum can be approximated in a data-driven way. Instead of only focusing on the abstract object state, Qi et al.~\yrcite{rpcin} propose Region Proposal Convolutional Interaction Network (RPCIN), a vision-based dynamics prediction model that takes raw image and descriptions of objects, such as bounding box and segmentation of objects which can be directly extracted from image, as input, so that it is more applicable for real-world scenario without instrumented setting~\cite{rpcin}. By adopting Region of Interest (RoI) Pooling~\cite{fastrcnn} and convolutional neural network (CNN), RPCIN incorporates visual information of both objects and environment, where the latter one tends to be ignored in previous works, and achieves state-of-the-art (SOTA) results, especially for long-term prediction. However, in spite of its generality to input requirements, as typical with DNNs, vision-based dynamics prediction models, like RPCIN, can be vulnerable to environment misalignment between the training and testing. In this paper, by using RPCIN as probe, we explore two types of environment misalignment challenges: \emph{Cross-Context} and \emph{Cross-Domain}, which can significantly compromise the model capability.

\emph{Cross-Domain} environment misalignment, where the visual domain is different between training and testing, challenges the model performance, where possessing such transferability can be critical. In the real-world, although it might be easy to gather static visual information of common objects, it can be difficult to collect dynamic visual information that capture the physics properties. For example, car images are easily acquirable whereas videos of car collisions are rare. While it is possible to synthesize such scenes, 
transferring a model trained on synthetic data to the real world leads to performance degradation. Chang et al.~\yrcite{compo_obj_base_phys} postulated that the visual appearance and the dynamic properties of the object are disentangled and should be separately modeled. This postulate, therefore, narrows the discussion to be on the state space of the objects, where the inputs are semantic properties of objects rather than images, while the visual environment characteristics are ignored~\cite{rpcin}. Thus, the actual real-world challenge of shifting the visual domain of the entire environment still remains under-explored.

\emph{Cross-Context} focuses on another aspect of environment misalignment challenge, where, even if the visual domain stays the same, the environment context is altered. For instance, a self-driving vehicle trained under normal traffic may suffer from irregular road condition, such as road closure with hazard sign. Most model generality discussions in previous works either ignore the changes to the environment context~\cite{ physics_play_billiard, compo_video_pred, rpcin} or represent the environment by the composition of designed objects~\cite{interaction_learn_phys, compo_obj_base_phys, phyre}, where the relationships between objects along with different physics properties of an object, such as position, mass, and velocity, may appear during training. Furthermore, efficiently and accurately decomposing an arbitrary environment by using pre-defined objects is also challenging. Thus, we seek to directly introduce alteration to the environment context, which is distinct from varying objects composition, between training and testing stage, so that the model has to be able to correctly derive environment context characteristic by only leveraging the raw visual information.

To the best of our knowledge, there does not exist dataset benchmarks that can be used to characterize the performance of vision-based long-term dynamics prediction models, in the presence of environment misalignment challenges. To investigate the \emph{cross-domain} and \emph{cross-context} challenges, a set of matching datasets is necessary, where, while the visual domain or environment context is different, the underlying dynamics stay same. Thus, we propose four datasets: \emph{SimB-Border}, \emph{SimB-Split}, \emph{BlenB-Border}, and \emph{BlenB-Split}, which cover two visual domains and two environment contexts.

We show that the performance of SOTA dynamics prediction method, e.g. RPCIN, significantly degrades on either \emph{Cross-Domain} or \emph{Cross-Context} challenge. Further, for addressing \emph{Cross-Domain} challenge, following Chang et al.~\yrcite{compo_obj_base_phys}, we extend their postulate and argue that we should seek a common intermediate representation space for both object and environment. Inspired by Bakhtin et al.~\yrcite{phyre}, where the data is simply provided as semantic masks, we argue that the semantic segmentation space can serve as the intermediate space, where a visual observation model~\cite{fcn, maskrcnn} first maps raw images to semantic segmentation masks, and then, the masks, along with the static information of objects, are used for predicting the dynamics. Our experiments show that, on the proposed dataset with \emph{Cross-Domain} setup, the performance by using ground-truth masks as input to RPCIN can significantly exceed the performance by using raw image as input. Even in the case that the ground-truth mask is absent, sub-optimal masks, which can be obtained via self-supervised learning, can also dramatically mitigate the \emph{Cross-Domain} challenge.

\input{sections/figures_text/cross_domain_figure.tex}

Our contributions can be summarised as follow: \footnote{Project is available at: \url{https://github.com/vimal-isi-edu/VDP-EMC}}
\begin{itemize}[noitemsep,topsep=0pt,parsep=0pt,partopsep=0pt]
    \item We identify two environment misalignment challenges, \emph{Cross-Domain} and \emph{Cross-Context}, for vision-based long-term dynamics prediction by using RPCIN as a probe.
    \item We propose four datasets, \emph{SimB-Border}, \emph{SimB-Split}, \emph{BlenB-Border}, and \emph{BlenB-Split} that cover two visual domains and two environment contexts for assessing the model performance on the challenges. 
    \item We discuss a promising direction for mitigating \emph{Cross-Domain} challenge and provide an intuitive instance as a concretization.
\end{itemize}

%% file: sections/figures_text/concept_figure.tex
\begin{figure} 
    \centering
     \includegraphics[width=\linewidth]{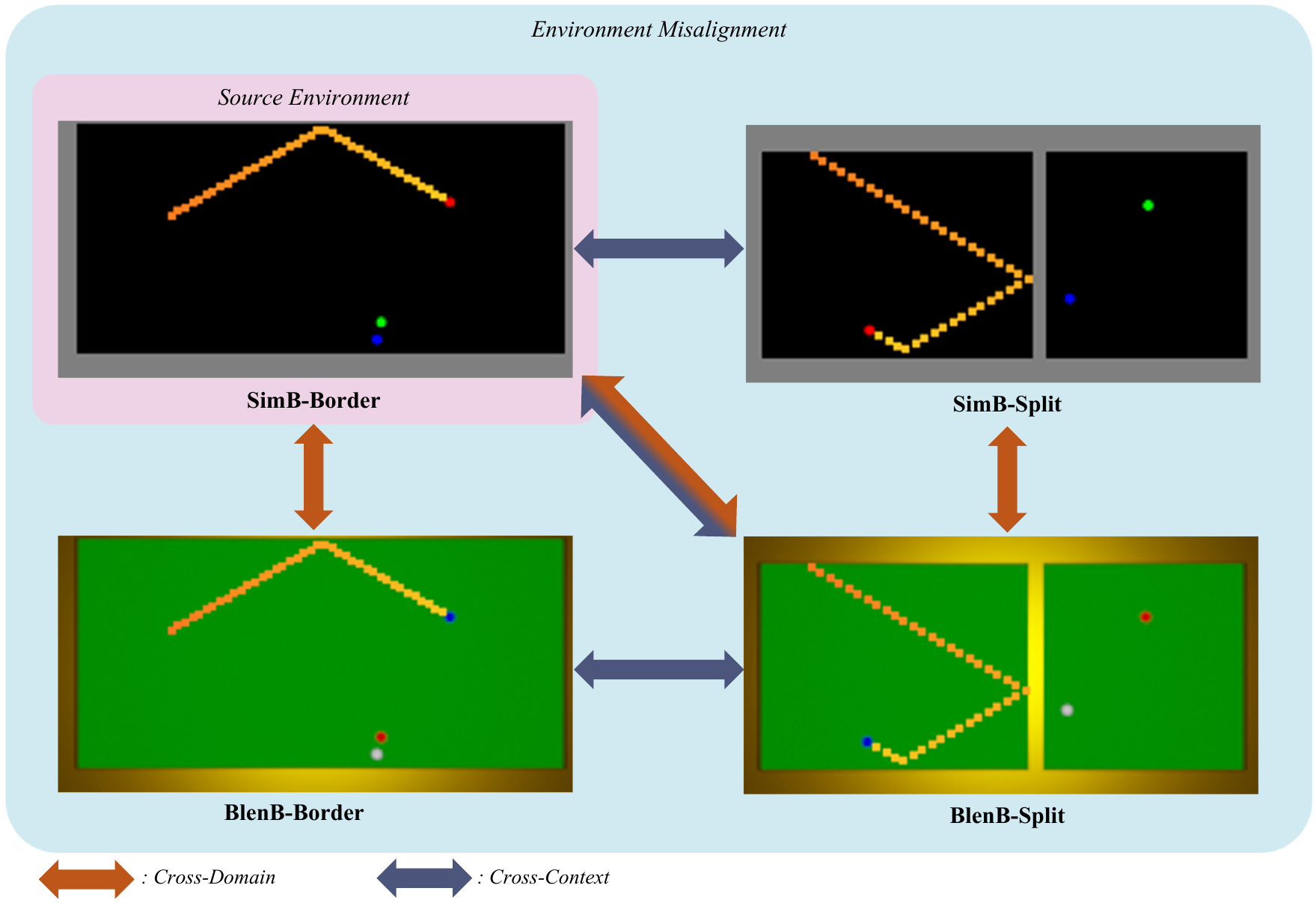}
    \caption{Illustration of two Environment Misalignment challenges: \emph{Cross-Domain} and \emph{Cross-Context}, and samples of four proposed datasets: \emph{SimB-Border}, \emph{SimB-Split}, \emph{BlenB-Border}, and \emph{BlenB-Split}. We adjusted images resolution and only display the ground-truth paths of the ball with displacement over time for better visualization. 
    } \label{fig:concept}
\end{figure}

%% file: sections/figures_text/cross_domain_figure.tex
\begin{figure*} 
    \centering
     \includegraphics[width=0.99\textwidth]{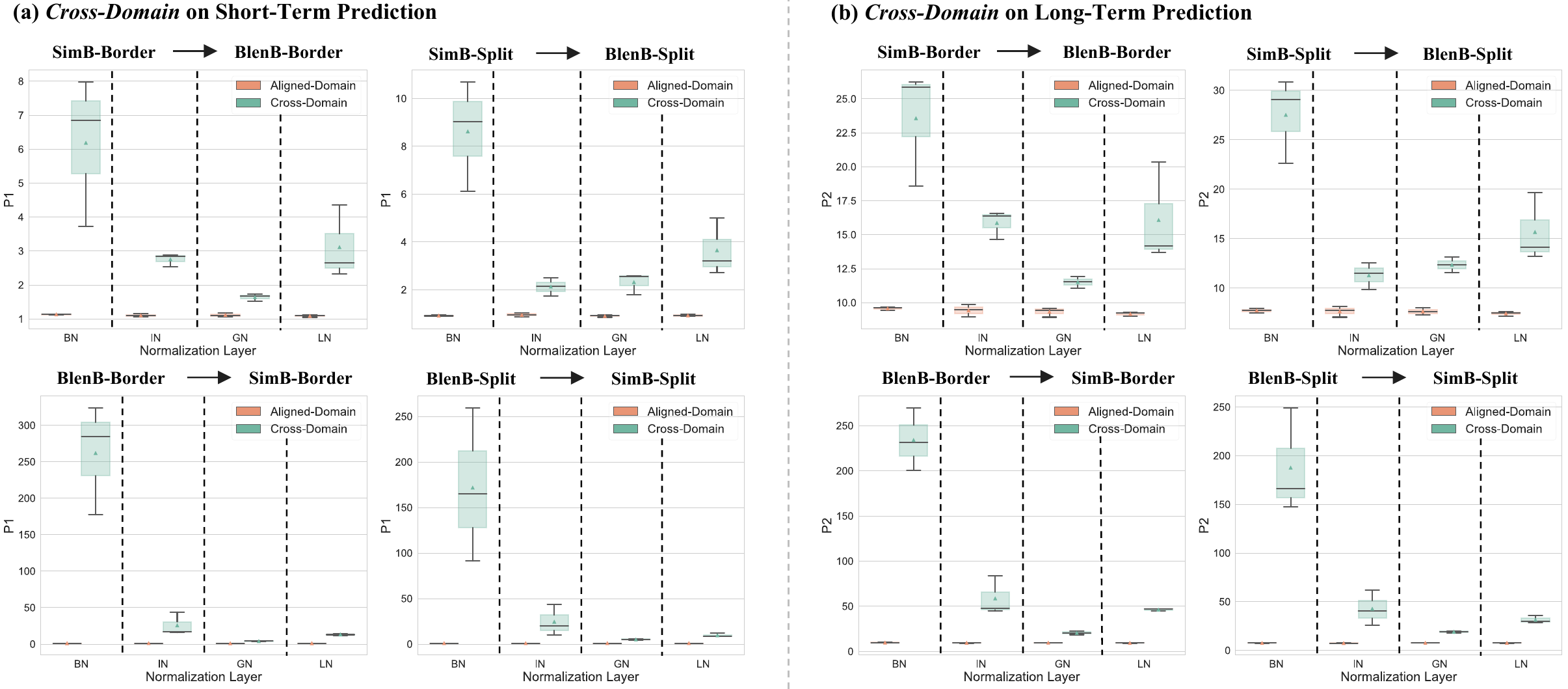}
    \caption{Performance of RPCIN with different normalization layers on \emph{Cross-Domain} challenge. Numerical results are in \cref{tab:sim_to_blen,tab:blen_to_sim}.} \label{fig:cross_domain}
\end{figure*}

%% file: sections/02_related_works.tex
\section{Related Works}
\textbf{Dynamics Prediction} is proposed as an instance of learning physics where the model is trained to derive the future states of objects or the scene given reference states as input~\cite{interaction_learn_phys, compo_obj_base_phys, compo_video_pred, reasoing_phys_interaction, learn_simulate_phys, rpcin}, where the dynamics model can be further applied to various tasks, such as dynamics planning~\cite{phyre, rpcin, mechanism_phy_reason} and dynamics visual reasoning~\cite{clevrer, dynamic_reason_diff_phys, dynamic_visual_reason}. Compared to considering the entire input scene as a whole, object-centric approaches~\cite{physics_play_billiard, compo_obj_base_phys, compo_video_pred} focus on each object of interest individually, where the relations between objects can be derived using interaction networks~\cite{interaction_learn_phys, rpcin}. Instead of relying on the manually defined physics models or requiring sophisticated object physics properties as input to infer dynamics~\cite{mujuco, ullman2014learning, compo_obj_base_phys, visual_de_animation, ete_diff_phy, dynamic_reason_diff_phys}, Qi et al.~\yrcite{rpcin} propose RPCIN for long-term dynamics prediction by only consuming raw images and simple object descriptions, such as bounding boxes and segmentation masks, as input, which is more feasible for real-world scenarios, and achieves SOTA performance.

\textbf{Domain Adaptation} has been studied on various vision tasks~\cite{ domain_adapt_fastrcnn, domainnet, advent} for adapting the trained model from source visual domain to target visual domain. Literature has been created on learning to adapt with multiple supervision signal strengths~\cite{unsupervised_da, semi_supervised_da, weakly_supervised_da}, between visual domains that share various degrees of category overlapping~\cite{partial_da, universal_da}, and with source-free setting where source domain data is unavailable for adaptation~\cite{source_free_adapt, source_free_adapt_segmentation}. Similar with source-free setting, for better applying to nowadays real-world scenarios where the source data can be inaccessible~\cite{source_free_adapt}, we also preclude the trained model from seeing the source data during adaptation. In conventional domain adaptation setting, where, despite vision tasks and data availability varying, the visual data itself in the target domain is sufficient for task-specific learning. 
Contrarily, in our proposed \emph{Cross-Domain} challenge, model is limited from accessing the information that the physics mechanism is inherent in, such as temporal sequence. Thus, directly applying the conventional domain adaptation methods to the identified challenge may 
be impractical. 

%% file: sections/03_method.tex
\input{sections/figures_text/cross_context_figure.tex}

\section{Method}
In this section, we first introduce the problem setup of dynamics prediction and background about RPCIN, which achieves SOTA performance. Then, we propose four datasets that cover two visual domains and two environment contexts, followed by investigating \emph{Cross-Domain} and \emph{Cross-Context} challenges individually by using RPCIN as a case study. Finally, we discuss an encouraging direction for mitigating \emph{Cross-Domain} challenge by mapping  raw images to a common intermediate space prior to learning dynamics prediction, and discuss an intuitive instance of this scheme. 

\subsection{Task Formulation and Preliminary} \label{sec:task_form_pre}
In this paper, we focus on the dynamics prediction in billiard game scenario where the evaluations are conducted on both short-term and long-term prediction~\cite{rpcin}. During training, model takes ${T_{ref}}$ consecutive image frames $\{X_{1-T_{ref}} ... X_{0}\}$ and reference states which describe ball states in each frame $\{S_{1-T_{ref}} ... S_{0}\}$ as input and  predicts ball states of next $T_{pred}$ frames $\{S'_{1}...S'_{T_{pred}}\}$, where the ground truth states $\{S_{1}...S_{T_{pred}}\}$ are provided as a supervision signal. During inference, the model also consumes $\{X_{1-T_{ref}} ... X_{0}\}$ and $\{S_{1-T_{ref}} ... S_{0}\}$ for predicting $\{S'_{1}...S'_{T_{pred}}\}$ and $\{S'_{T_{pred}+1}...S'_{2T_{pred}}\}$ which are evaluated as short-term and long-term predictions, respectively.

RPCIN~\cite{rpcin} proposes to solve this problem end-to-end by extending the interaction network~\cite{interaction_learn_phys}, and only requiring the bounding box of each ball to represent the frame state $S$. By utilizing RoI Pooling~\cite{fastrcnn}, for each reference frame, each ball state feature $b_i$ can be directly extracted from the visual feature which is encoded from the raw image through a DNN. For completeness, we will briefly summarize RPCIN here and refer the readers to Qi et al.~\yrcite{rpcin} for details. For inferring the dynamics of each ball, borrowing the notations from Qi et al. ~\yrcite{rpcin}, RPCIN defines five CNNs. $f_O$ takes the state feature $b_i^t$ of the $i$-th ball at the $t$-th timestep as input to infer its self-dynamic features. $f_R$ pair-wisely takes $b_i^t$ and each of the state features $b_j^t$ of other balls at the $t$-th timestep as input to infer the pair-wise relative-dynamics, where the summation combines them for the total relative-dynamics. $f_A$ takes the sum of self- and relative-dynamics features for systematically inferring the overall-dynamics features, where the results, along with the state feature $b_i^t$, are consumed by $f_Z$ for a static-dynamics mixture feature. Finally, $f_P$ takes the mixture features of the previous $T_{ref}$ time steps to predict the state feature of the the $i$-th ball at the next time step. The entire process is shown in \cref{eq:rpcin_model} (Qi et al.~\yrcite{rpcin}):
\begin{equation}
\label{eq:rpcin_model}
\begin{aligned}
    e_i^t &= f_A(f_O(b_i^t) + \sum_{j \neq i}f_R(b_i^t, b_j^t)), \\
    z_i^t &= f_Z(b_i^t, e_i^t), \\
    b_i^{t+1} &= f_P(z_i^t, z_i^{t-1},...,z_i^{t-T_{ref}+1})
\end{aligned}
\end{equation}

\input{sections/figures_text/visualization_figure.tex}

\input{tables/dataset_details}

\subsection{Datasets for Environment Misalignments} \label{sec:dataset}
\emph{SimB} dataset was used in RPCIN, which simulates the movements of and collisions of three balls without any information about the context of the environment, besides the image boundaries~\cite{rpcin}. Thus, we find this dataset not suitable for our usage due to its over simplicity. To this end, we propose \emph{SimB-Border} as an extension of \emph{SimB} by increasing the image size from $64\times64$ to $192\times96$, which includes more space for adding various contexts of environment, and introducing borders to the image boundaries. The width of the borders for the four boundaries are individually and randomly selected as integers in the range of $[0, 15]$, where zero stands for no border, and kept consistent for all frames in a single video. Introducing borders is important since it challenges the model to correctly extract the border properties from the raw image to precisely predict the locations where the ball bounces back. Additionally, to further increase the prediction difficulty and penalize the model for failing to understand the visual context, we extend \emph{SimB-Border} to \emph{SimB-Split} by adding a five-pixels wide vertical bar into the scene. For each video, the horizontal center of the bar is randomly selected as an integer in the range of [64, 128] and kept consistent over the video frames.

Aside from the \emph{Sim} domain, for investigating the performance of the model under dissimilar visual domains, it is essential to create matching datasets in a different domain with closely comparable underlying dynamics. Empowered by having access to the full ground-truth environment contexts and objects information of datasets in \emph{Sim} domain, we use Blender~\cite{blender} to create matching datasets, \emph{BlenB-Border} and \emph{BlenB-Split}, in \emph{Blen} domain. In Blender, borders, split, and balls are rendered from objects where their properties, such as width, length, and locations, are adjusted by taking the ground-truth information of samples in \emph{Sim} domain as guidance. Since Blender takes metric units of length (e.g., meter or inch) as dimension measurement, where the sizes in pixel in the rendered images are relative values controlled by multiple Blender parameters, we manually adjusted relevant factors for seeking a close match between \emph{Sim} and \emph{Blen} domains. More details about creating dataset in \emph{Blen} domain are included in the Appendix. Visualizations of the four proposed datasets are shown in \cref{fig:concept} and details are include in \cref{tab:data_detail}.

\input{tables/sim_to_blen_table.tex}

\subsection{Cross-Domain Misalignment} \label{sec:cross_domain}
For the \emph{Cross-Domain} challenge, we extend the problem setup described in \cref{sec:task_form_pre} with two visual domains: Source Domain and Target Domain. During training, data on the source domain contains all information including image frames, static information of balls, and temporal sequence that connect individual frames for learning dynamics. On the target domain, the model still has access to individual frames and the static balls information, but the temporal sequence, where the physical dynamics are evident, is absent so that learning correct dynamics prediction is impractical. Furthermore, to increase the model's generality in the real-world, where the source domain data may be unavailable after domain shifts~\cite{source_free_adapt}, we also limit the trained model from accessing the source domain data. The setup of testing data on both domains is the same as described in \cref{sec:task_form_pre}. Since the only substantial difference between datasets in \emph{Sim} and \emph{Blen} domains is the visual appearance, given an optimal appearance-agnostic dynamics prediction model, the performance should be comparable across visual domains. Due to the lack of sufficient information for fine tuning RPCIN model on the target domain, we directly evaluate the source domain trained model on the target domain. Furthermore, considering prior works in the domain adaptation field~\cite{revisitbn, domainsepcibn} showing that various normalization methods in DNN may significantly affect the performance of a model when visual domain shifts occur, we conduct experiments with several normalization methods to comprehensively validate the performance of RPCIN under environment misalignment challenges. In detail, in addition to the widely used Batch Normalization (BN)~\cite{batchnorm}, we also explore Instance Normalization (IN)~\cite{instnorm}, Layer Normalization (LN)~\cite{layernorm}, and Group Normalization (GN)~\cite{groupnorm}.

As shown in \cref{fig:cross_domain}, the model performance significantly decreases when the domain changes, where the impacts are more severe for applying the models trained on \emph{Blen} domain to \emph{Sim} domain. This might be due to the fact that, compared to extracting information from over simplified visual appearance of \emph{Sim} domain, models are more likely to overfit to \emph{Blen} domain so that the generality is further compromised. Furthermore, it is also noticeable that the model with BN, which are arguably the most commonly used architecture, is most vulnerable. This is due to the heavy dependence of its normalization statistics on the domain specific knowledge~\cite{revisitbn}. Unlike the conventional domain adaptation problem, where the visual data that contains sufficient task-specific information is available, such information specific for dynamics prediction is unavailable for the target domain in the proposed \emph{Cross-Domain} challenge because the correct temporal sequence, where the underlying mechanism is inherent in, is missing. For instance, correctly inferring the dynamics of a ball is infeasible with merely one image of the ball that only contains the static properties of the ball rather than a sequence of images that can describe the movement of the ball.

\input{tables/blen_to_sim_table.tex}

\subsection{Cross-Context Misalignment} \label{sec:cross_context}
Similar to the setup of \emph{Cross-Domain} challenge, as described in ~\cref{sec:cross_domain}, we also consider two environment contexts for \emph{Cross-Context} challenge: Source Context and Target Context, where, during training, in source context, the model has access to all information, as described in \cref{sec:task_form_pre}. After shifting to the target context, correct temporal sequence and source context data become unavailable. The setup of testing data on both context is the same as described in \cref{sec:task_form_pre}.  Unlike the model generality discussions in previous works, our proposed \emph{Cross-Context} challenge focuses on altering the environment context, which is not represented by any object whose static information is explicitly provided for dynamics prediction. As shown in ~\cref{fig:cross_context}, even when the visual domains are exactly the same, the performance of the model dramatically decreases by simply placing or removing a split bar in the middle of the environment context. As visualizations shown in ~\cref{fig:visual}(a), for the models trained in the \emph{Split} context, they tend to hallucinate a split bar in the middle of the image, whereas for the model trained in the \emph{Border} context, they tend to ignore the  bar. We hypothesize that such behavior might be because, in addition to predicting the state feature that only encodes object static information in the future, the model also tends to fuse auxiliary knowledge with respect to the particular environment context as a short-cut to minimize the empirical loss. For instance, in \emph{Split} context, besides the visual information of the split bar, model may also be counting the prediction steps for determining the bouncing back location. Thus, in \emph{Border} context, such overfitting can mislead the model for wrong predictions.

\subsection{Unifying Visual Domains with Segmentation Masks} \label{sec:visual_mask}
One of the core struggles that limits the model adaptation in the \emph{Cross-Domain} challenge for vision-based dynamics prediction model is the absence of object dynamics information in the target domain, such as the correct temporal sequence as described in \Cref{sec:cross_domain}. However, the data for extracting the static information is available on both source and target domains. Therefore, the correct dynamics prediction is expected to be made on the target domain by feeding the aligned static information into the pretrained dynamics model, along with the proper reference temporal sequence once they become available during test. However, since the vision-based dynamics prediction model may entangle visual and dynamics information, such static information alignment can be difficult to reach without concurrently accessing the data from both domains, which is prohibited in our setup. Thus, we extend the belief held by Chang et al.~\yrcite{compo_obj_base_phys} to that the entire visual scene, including both objects and environment context, shall be mapped to an intermediate common abstraction space before feeding to the dynamics prediction model. In the scope of the billiard datasets discussed in this paper, since all balls being assumed to share the same physics properties, such as mass, radius, and shape, the bounding box of each ball is a sufficient object abstraction for dynamics prediction.

To seek an instance of environment context abstraction, inspired by PHYRE~\cite{phyre}, where the data is directly provided as semantic masks, we argue that the semantic segmentation map can serve as an adequate abstraction. With successes on the visual observation tasks, acquiring the segmentation maps, even in the conventional domain adaptation~\cite{advent} or unsupervised setting~\cite{iic}, is feasible. To investigate the feasibility resolving \emph{Cross-Domain} challenge by replacing RGB image with semantic segmentation mask as input to the dynamics prediction model, we considered the mask obtained from three sources: ground-truth mask (GT-Mask), supervised trained mask (Sup-Mask), and self-supervised trained mask (Self-Mask). For the environment context in the proposed dataset, there are two semantic classes: table, where the ball traverses, and border, including split bar, which bounds the movement of the balls. Thus, the GT-Mask is a two-channel mask, and is used as supervision signal to train the semantic segmentation model which is used to extract the Sup-Mask. For relaxing the requirement of having access to the ground-truth masks, which might be hard to gather in the real-world, we also train the segmentation model with self-supervised learning. Inspired by the method proposed by Caron et al.~\yrcite{kmean_pseudo}, we use k-means to extract masks for training images which serves as the supervision signal for training the segmentation model. Visualization of various masks are shown in \cref{fig:visual}(b) and the training details are included in Appendix. Experiments show that, on the proposed datasets, comparing with baseline results of various normalization methods, using GT-Mask as replacement can principally resolve the \emph{Cross-Domain} challenge, and even using Self-Mask as replacement can achieve comparable results. Detail discussions are included in \cref{sec:dis_cross_domain}.

%% file: sections/figures_text/cross_context_figure.tex
\begin{figure*} 
    \centering
     \includegraphics[width=0.99\textwidth]{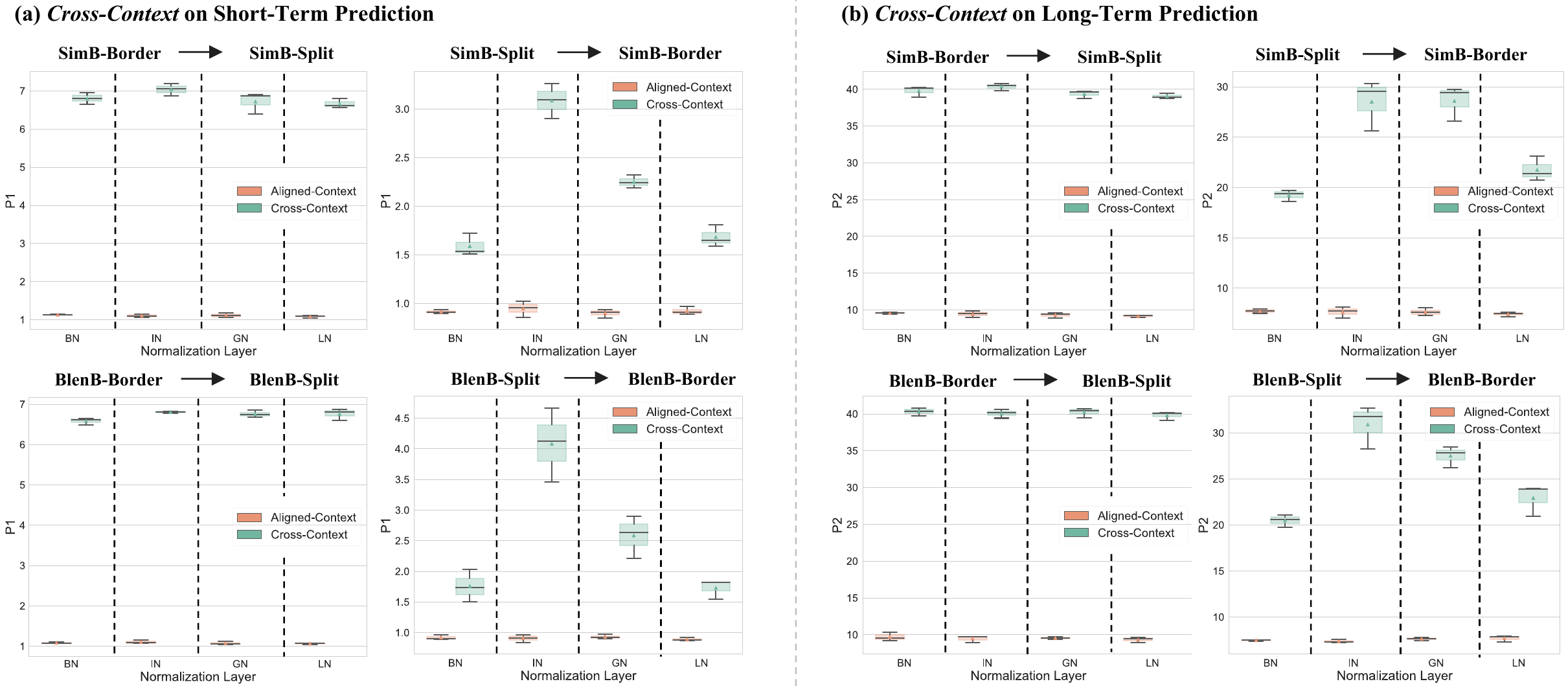}
    \caption{Performance of RPCIN with different normalization layers on \emph{Cross-Context} challenge. Numerical results are in Appendix.} \label{fig:cross_context}
\end{figure*}

%% file: sections/figures_text/visualization_figure.tex
\begin{figure*} 
    \centering
     \includegraphics[width=\textwidth]{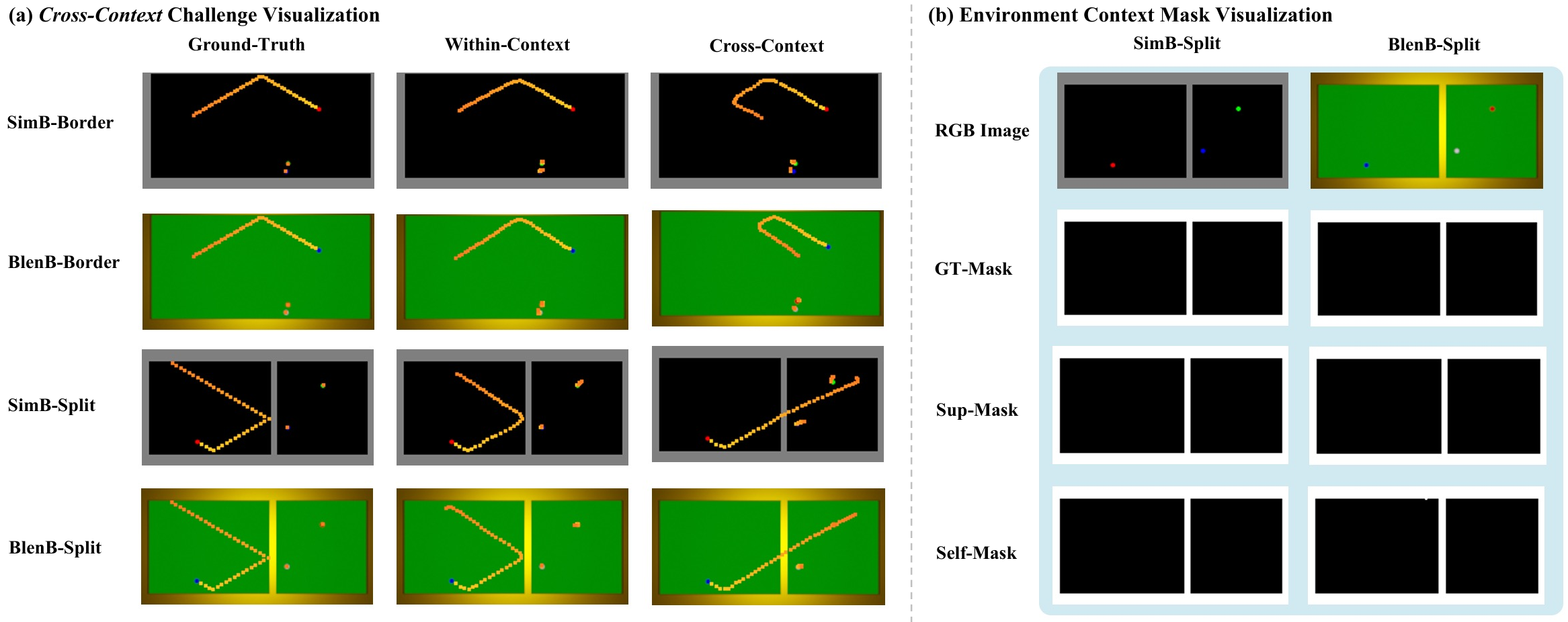}
    \caption{Visualizations of (a) \emph{Cross-Context challenge} where trained model fails on the target context, and (b) various environment context mask. We adjusted images resolution for better visualization.} \label{fig:visual}
\end{figure*}

%% file: tables/dataset_details.tex
\begin{table}[htp]
\centering
\caption{Details of the four proposed datasets and the original SimB dataset. \textit{Sim} stands for the visual domain of simple simulation and \textit{Blen} stands for the visual domain rendered by Blender.} \label{tab:data_detail}
\begin{adjustbox}{width=\linewidth}
\renewcommand{\arraystretch}{1.3}
\begin{tabular}{c|ccccc}
\hlineB{3}
Dataset      & Resolution   & Domain     & Context & Border Width & Split Center \\ \hline
SimB~\cite{rpcin}         & 64 $\times$ 64  & \textit{Sim} & N/A     & N/A         & N/A            \\
SimB-Border  & 192 $\times$ 96 & \textit{Sim} & Border  & {[}0, 15{]} & N/A            \\
SimB-Split   & 192 $\times$ 96 & \textit{Sim} & Split   & {[}0, 15{]} & {[}64, 128{]}  \\
BlenB-Border & 192 $\times$ 96 & \textit{Blen}    & Border  & {[}0, 15{]} & N/A            \\
BlenB-Border & 192 $\times$ 96 & \textit{Blen}    & Split   & {[}0, 15{]} & {[}64, 128{]}  \\ \hlineB{3}
\end{tabular}
\end{adjustbox}
\end{table}

%% file: tables/sim_to_blen_table.tex
\begin{table*}[htp]
\centering
\caption{Performance of \emph{Sim} domain trained RPCIN model on \emph{Cross-Domain} challenge with various types of input. BN is used for all segmentation mask input training. In addition to the BN baseline, which takes RGB image as input, we also list other normalization method results as baselines for a comprehensive comparison and further demonstrating the advantage of unifying the visual domains with segmentation masks. Aligned and Cross results are the same for GT-Mask trained model because they share exactly the same data. \textbf{Bold} highlights the best results and \underline{underline} highlights the second best results.} \label{tab:sim_to_blen}
\begin{adjustbox}{width=0.99\textwidth}
\renewcommand{\arraystretch}{1.3}
\begin{tabular}{ccccccccc}
\hlineB{3}
Source Dataset & \multicolumn{4}{c|}{SimB-Border}                                                                              & \multicolumn{4}{c}{SimB-Split}                                                                  \\ \cline{2-9} 
Target Dataset & \multicolumn{2}{c|}{SimB-Border (\red{Aligned})}            & \multicolumn{2}{c|}{BlenB-Border (\red{Cross})}              & \multicolumn{2}{c|}{SimB-Split (\red{Aligned})}             & \multicolumn{2}{c}{BlenB-Split (\red{Cross})} \\
Eval Period    & P1                                 & \multicolumn{1}{c|}{P2}                                & P1                                & \multicolumn{1}{c|}{P2}                                & P1                                & \multicolumn{1}{c|}{P2}                                & P1                                         & P2                                          \\ \hline
\multicolumn{9}{c}{Raw RGB Image Input}                                                                                                                                                                                                                                                                                                                                                           \\ \hline
RGB-BN         & 1.131 $\pm$ 0.011                     & \multicolumn{1}{c|}{9.568 $\pm$ 0.121}                    & 6.185 $\pm$ 2.206                    & \multicolumn{1}{c|}{23.564 $\pm$ 4.307}                   & 0.913 $\pm$ 0.019                    & \multicolumn{1}{c|}{7.732 $\pm$ 0.208}                    & 8.622 $\pm$ 2.313                             & 27.511 $\pm$ 4.322                             \\
RGB-IN         & 1.102 $\pm$ 0.045                     & \multicolumn{1}{c|}{9.426 $\pm$ 0.446}                    & 2.754 $\pm$ 0.188                    & \multicolumn{1}{c|}{15.863 $\pm$ 1.073}                   & 0.945 $\pm$ 0.082                    & \multicolumn{1}{c|}{7.641 $\pm$ 0.549}                    & 2.127 $\pm$ 0.381                             & 11.281 $\pm$ 1.358                             \\
RGB-GN         & 1.117 $\pm$ 0.058                     & \multicolumn{1}{c|}{\underline{9.323 $\pm$ 0.346}}                    & 1.637 $\pm$ 0.106                    & \multicolumn{1}{c|}{11.507 $\pm$ 0.425}                   & \textbf{0.899 $\pm$ 0.042}                    & \multicolumn{1}{c|}{7.632 $\pm$ 0.386}                    & 2.315 $\pm$ 0.447                             & 12.335 $\pm$ 0.792                             \\
RGB-LN         & \textbf{1.085 $\pm$ 0.033}                     & \multicolumn{1}{c|}{\textbf{9.165 $\pm$ 0.145}}                    & 3.114 $\pm$1.086                     & \multicolumn{1}{c|}{16.074 $\pm$ 3.716}                   & 0.922 $\pm$ 0.042                    & \multicolumn{1}{c|}{7.433 $\pm$ 0.237}                   & 3.648 $\pm$ 1.199                             & 15.662 $\pm$ 3.483                             \\ \hline
\multicolumn{9}{c}{Segmentation Mask Input}                                                                                                                                                                                                                                                                                                                                                       \\ \hline
GT-Mask        & \underline{1.091 $\pm$ 0.044}                     & \multicolumn{1}{c|}{9.358 $\pm$ 0.465}                    & \textbf{1.091 $\pm$ 0.044}                    & \multicolumn{1}{c|}{\textbf{9.358 $\pm$ 0.465}}                    & 0.916 $\pm$ 0.005                    & \multicolumn{1}{c|}{\underline{7.431 $\pm$ 0.511}}                    & \textbf{0.916 $\pm$ 0.005}                             & \underline{7.431 $\pm$ 0.511}                              \\
Sup-Mask       & 1.093 $\pm$ 0.021                     & \multicolumn{1}{c|}{9.396 $\pm$ 0.285}                    & \underline{1.093 $\pm$ 0.021}                    & \multicolumn{1}{c|}{\underline{9.397 $\pm$ 0.286}}                    & 0.971 $\pm$ 0.011                    & \multicolumn{1}{c|}{\textbf{7.372 $\pm$ 0.089}}                    & 0.981 $\pm$ 0.012                             & \textbf{7.422 $\pm$ 0.095}                              \\
Self-Mask     & 1.119 $\pm$ 0.037                     & \multicolumn{1}{c|}{9.604 $\pm$ 0.300}                    & 1.132 $\pm$ 0.035                    & \multicolumn{1}{c|}{9.614 $\pm$ 0.291}                    & \underline{0.911 $\pm$ 0.025}                    & \multicolumn{1}{c|}{7.837 $\pm$ 1.334}                    & \underline{0.959 $\pm$ 0.020}                             & 8.017 $\pm$ 1.309                              \\ \hlineB{3}
\end{tabular}
\end{adjustbox}
\end{table*}

%% file: tables/blen_to_sim_table.tex
\begin{table*}[htp]
\centering
\caption{Performance of \emph{Blen} domain trained RPCIN model on \emph{Cross-Domain} challenge with various types of input. BN is used for all segmentation mask input training. In addition to the BN baseline, which takes RGB image as input, we also list other normalization method results as baselines for a comprehensive comparison and demonstrating the advantage of unifying visual domains with segmentation masks. Aligned and Cross results are the same with ~\cref{tab:sim_to_blen} for GT-Mask trained model because they share exactly the same data. \textbf{Bold} highlights the best results and \underline{underline} highlights the second best results. } \label{tab:blen_to_sim}
\begin{adjustbox}{width=0.99\textwidth}
\renewcommand{\arraystretch}{1.3}
\begin{tabular}{ccccccccc}
\hlineB{3}
Source Dataset & \multicolumn{4}{c|}{BlenB-Border}                                                                                  & \multicolumn{4}{c}{BlenB-Split}                                                               \\ \cline{2-9} 
Target Dataset & \multicolumn{2}{c|}{BlenB-Border (\red{Aligned})}           & \multicolumn{2}{c|}{SimB-Border (\red{Cross})}                    & \multicolumn{2}{c|}{BlenB-Split (\red{Aligned})}           & \multicolumn{2}{c}{SimB-Split (\red{Cross})} \\ 
Eval Period    & P1             & \multicolumn{1}{c|}{P2}             & P1                & \multicolumn{1}{c|}{P2}                & P1             & \multicolumn{1}{c|}{P2}             & P1                 & P2                \\ \hline
\multicolumn{9}{c}{Raw RGB Image Input}                                                                                                                                                                                            \\ \hline
RGB-BN         & 1.084 $\pm$ 0.023 & \multicolumn{1}{c|}{9.713 $\pm$ 0.554} & 261.768 $\pm$ 75.655 & \multicolumn{1}{c|}{233.977 $\pm$ 34.460} & 0.918 $\pm$ 0.037 & \multicolumn{1}{c|}{7.478 $\pm$ 0.079} & 171.750 $\pm$ 84.274  & 187.635 $\pm$ 53.940 \\
RGB-IN         & 1.103 $\pm$ 0.041 & \multicolumn{1}{c|}{9.471 $\pm$ 0.443} & 25.600 $\pm$ 15.781  & \multicolumn{1}{c|}{58.599 $\pm$ 21.827}  & 0.906 $\pm$ 0.062 & \multicolumn{1}{c|}{\underline{7.368 $\pm$ 0.168}} & 24.460 $\pm$ 17.239   & 42.613 $\pm$ 18.159  \\
RGB-GN         & \underline{1.075 $\pm$ 0.045} & \multicolumn{1}{c|}{9.560 $\pm$ 0.145} & 3.899 $\pm$ 0.526    & \multicolumn{1}{c|}{20.425 $\pm$ 1.913}   & 0.931 $\pm$ 0.039 & \multicolumn{1}{c|}{7.641 $\pm$ 0.165} & 5.033 $\pm$ 0.575     & 18.970 $\pm$ 1.323   \\
RGB-LN         & \textbf{1.064 $\pm$ 0.020} & \multicolumn{1}{c|}{\textbf{9.345 $\pm$ 0.350}} & 12.969 $\pm$ 1.508   & \multicolumn{1}{c|}{46.113 $\pm$ 1.157}   & \underline{0.892 $\pm$ 0.027} & \multicolumn{1}{c|}{7.687 $\pm$ 0.321} & 9.518 $\pm$ 2.024     & 31.364 $\pm$ 3.740   \\ \hline
\multicolumn{9}{c}{Segmentation Mask Input}                                                                                                                                                                                        \\ \hline
GT-Mask        & 1.091 $\pm$ 0.044                     & \multicolumn{1}{c|}{9.358 $\pm$ 0.465}                    & \textbf{1.091 $\pm$ 0.044}                    & \multicolumn{1}{c|}{\underline{9.358 $\pm$ 0.465}}                    & 0.916 $\pm$ 0.005                    & \multicolumn{1}{c|}{7.431 $\pm$ 0.511}                    & \underline{0.916 $\pm$ 0.005}                             & \underline{7.431 $\pm$ 0.511}                              \\
Sup-Mask       & 1.122 $\pm$ 0.036 & \multicolumn{1}{c|}{\underline{9.353 $\pm$ 0.268}} & \underline{1.121 $\pm$ 0.036}    & \multicolumn{1}{c|}{\textbf{9.353 $\pm$ 0.268}}    & \textbf{0.891 $\pm$ 0.027} & \multicolumn{1}{c|}{\textbf{7.317 $\pm$ 0.273}} & \textbf{0.889 $\pm$ 0.027}     & \textbf{7.313 $\pm$ 0.272}    \\
Self-Mask     & 1.136 $\pm$ 0.024 & \multicolumn{1}{c|}{9.945 $\pm$ 0.563} & 1.136 $\pm$ 0.024    & \multicolumn{1}{c|}{9.943 $\pm$ 0.560}    & 0.914 $\pm$ 0.022 & \multicolumn{1}{c|}{7.539 $\pm$ 0.227} & 0.944 $\pm$ 0.023     & 7.650 $\pm$ 0.224    \\ \hlineB{3}
\end{tabular}
\end{adjustbox}
\end{table*}

%% file: sections/04_experiments.tex
\section{Experiments}
In this section, we first briefly introduce the experiment details, and then provide analysis on results of \emph{Cross-Domain} and \emph{Cross-Context} challenges. Finally, we discuss the limitations and open issues.

\input{sections/figures_text/ablation_figure.tex}

\subsection{Training and Evaluation Details}
We conduct experiments on the proposed \emph{SimB-Border}, \emph{SimB-Split}, \emph{BlenB-Border}, and \emph{BlenB-Split} datasets by using RPCIN~\cite{rpcin} as a baseline model, and its public implementation as our code base. Following RPCIN, we use Hourglass~\cite{hourglass} as the visual backbone, randomly flip the image horizontally and vertically as augmentation, and use the discounted loss~\cite{visual_interaction} for stabilizing the training. To comprehensively evaluate the model performance, as described in \cref{sec:cross_domain}, we replace the BN layer with GN, IN, and LN layer, where the group number in GN is set to 32. Since BN is arguably the most commonly used normalization layer, we use BN for all the training with masks as input. The RoI pooling output size is set to three. More details can be found in Appendix.

For experiment details, we set the length of reference frames ${T_{ref}}$ to four and training prediction period $T_{pred}$ to 20. Evaluations are separately conduced on short-term predictions $\{1...T_{pred}\}$ (P1) and long-term predictions $\{T_{pred}+1...2 \times T_{pred)}\}$ (P2), where the squared $l_2$ distance between the ground-truth and predicted objects bounding box center is used as the evaluation metric. The distance is further scaled by 1000 for better demonstration. All those settings are aligned with RPCIN~\cite{rpcin}.

\input{tables/mask_general.tex}

\subsection{Discussions on \emph{Cross-Domain}} \label{sec:dis_cross_domain}
\textbf{Segmentation mask as common intermediate abstract space: }In \cref{sec:visual_mask}, we argued for mitigating \emph{Cross-Domain} challenge by mapping the raw image to a common intermediate abstract space prior to dyanmics prediction. We also argued that the segmentation mask can serve as a promising abstract space. We conduct experiments by replacing the raw images with masks and the results are shown in \cref{tab:sim_to_blen,tab:blen_to_sim}. The masks for training and testing on source and target domain are the same kind. It is noticeable that all three kinds of mask, including Self-Mask, can dramatically mitigate the \emph{Cross-Domain} challenge and achieve a competitive performance compared with the best domain aligned results with raw images as input. Such outstanding performance empirically demonstrates the feasibility and strength of utilizing the segmentation mask as the common intermediate abstract space. 

\textbf{Generality between various kinds of mask: }As shown in \cref{fig:visual}(b), Sup-Mask and Self-Mask of both \emph{Sim} and \emph{Blen} domain are comparable with GT-Mask. Thus we further study a generality problem that the masks used for training and testing on the source and target domains are obtained via different supervision strength, which further mimic the real-world scenario where the source domain usually contains richer information than the target domain. As shown in \cref{tab:mask_gen}, even though the mask obtained with weaker supervision strength may be sub-optimal, the dynamics prediction model trained on the superior mask still preserves the robustness to the \emph{Cross-Domain} challenge. 

\subsection{Discussion on \emph{Cross-Context}}
\textbf{Alignment loss for the predicted state feature: }As described in \cref{sec:task_form_pre}, RPCIN extracts the reference ball state features $b^t_i$ from the reference image and recursively used it to predict the ball state features $b^{t+1}_i$ in the future. The latter one is further decoded to semantic output, such as bounding box. By replacing the raw images input with masks, future ball state features can also be extracted by utilizing the future ground-truth bounding box and reference mask, where we annotate this feature as $\hat{b}^{t+1}_i$. An intuitive path to constrain the dynamics model from encoding static irrelevant information to $b^{t+1}$ is by introducing an alignment loss which reduces the difference between $b^{t+1}_i$ and $\hat{b}^{t+1}_i$. We use mean-square loss as regularization with different loss weight and conduct experiments on GT-Mask. As results shown in \cref{fig:ablation}(a), although it seems that increasing the alignment loss weight may improve \emph{Cross-Context} endurance on \emph{SimB-Split} to \emph{SimB-Border}, it does not fundamentally address the challenge and the performance gaps are still huge. 

\textbf{Varying RoI Pooling Output Size: }The RoI pooling operation in RPCIN enables the model to consider the environment for dynamics prediction~\cite{rpcin}. By increasing the size of RoI pooling output size, richer information regarding the environment may be encoded to the ball state features that may help with enduring \emph{Cross-Context} challenge. However, as our results on GT-Mask shown in \cref{fig:ablation}(b), varying RoI pooling output size does not solve the \emph{Cross-Context} challenge.

\subsection{Limitations and Open Issues}
Despite the success of utilizing segmentation masks to mitigate the \emph{Cross-Domain} challenge, the \emph{Cross-Context} challenge is still outstanding. Even though we specifically disentangle the visual information from the dynamics information, the learned dynamics may still be entangled with other confounders, for example encoding context specific knowledge as a short-cut to minimize empirical loss. We hypothesis that, further disentanglement between underlying dynamics  and latent confounders is needed to address the \emph{Cross-Context} challenge. Further, our experiments assume all objects share the same physics properties and focus on the environment differences. However, combining  alterations to both environment and object properties can further push the boundary of model generality. Finally, our experiments are conducted using synthetic data in order to obtain consistent underlying mechanisms, which already challenges the vision-based dynamics prediction model. Creating a dataset of matching synthetic and real domains is still an open issue. Although we are aware of mechanism differences, we still include experimental results with RealB~\cite{rpcin} in Appendix for completion.

%% file: sections/figures_text/ablation_figure.tex
\begin{figure*} 
    \centering
     \includegraphics[width=\textwidth]{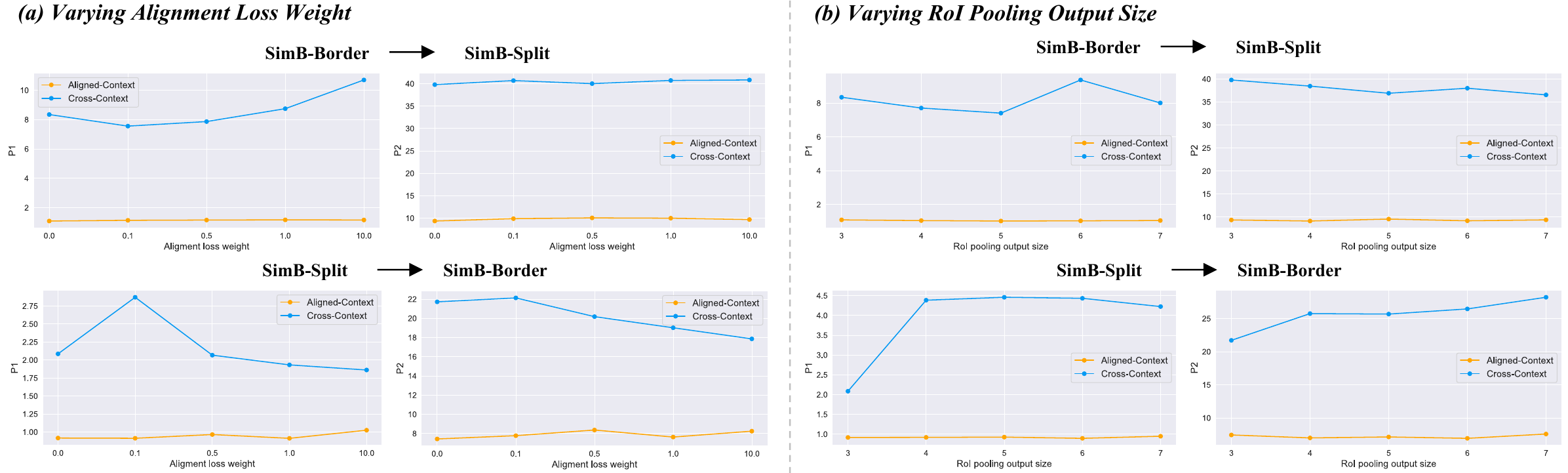}
    \caption{Ablation studies with GT-Masks as input on (a) varying alignment loss weight and (b) varying RoI pooling output size. 
    } \label{fig:ablation}
\end{figure*}

%% file: tables/mask_general.tex
\begin{table}[htp]
\centering
\caption{Study of generality between various kinds of mask. Masks for training and testing on the source and target domain are obtaiend via different supervision strength. For the same environment context, GT-Masks are the same between different visual domains.} \label{tab:mask_gen}
\begin{adjustbox}{width=\linewidth}
\renewcommand{\arraystretch}{1.3}
\begin{tabular}{ccccc}
\hlineB{3}
\multicolumn{5}{c}{Sim Domain $\rightarrow$ Blen Domain}                                                                                              \\ \hline
                                 & \multicolumn{2}{c|}{SimB-Border $\rightarrow$ BlenB-Border} & \multicolumn{2}{c}{SimB-Split $\rightarrow$ BlenB-Split} \\
                                 & P1               & \multicolumn{1}{c|}{P2}                & P1                          & P2                         \\ \hline
GT-Mask $\rightarrow$ GT-Mask    & 1.091 $\pm$ 0.044   & \multicolumn{1}{c|}{9.358 $\pm$ 0.465}    & 0.916 $\pm$ 0.005              & 7.431 $\pm$ 0.511             \\
GT-Mask $\rightarrow$ Sup-Mask   & 1.091 $\pm$ 0.044   & \multicolumn{1}{c|}{9.360 $\pm$ 0.463}    & 0.926 $\pm$ 0.008              & 7.479 $\pm$ 0.494             \\
GT-Mask $\rightarrow$ Self-Mask  & 1.280 $\pm$ 0.065   & \multicolumn{1}{c|}{10.105 $\pm$ 0.381}   & 1.236 $\pm$ 0.034              & 8.576 $\pm$ 0.399             \\
Sup-Mask $\rightarrow$ Self-Mask & 1.220 $\pm$ 0.035   & \multicolumn{1}{c|}{9.776 $\pm$ 0.323}    & 1.277 $\pm$ 0.070              & 8.446 $\pm$ 0.299             \\ \hline
\multicolumn{5}{c}{Blen Domain $\rightarrow$ Sim Domain}                                                                                              \\ \hline
                                 & \multicolumn{2}{c|}{BlenB-Border $\rightarrow$ SimB-Border} & \multicolumn{2}{c}{BlenB-Split $\rightarrow$ SimB-Split} \\
                                 & P1               & \multicolumn{1}{c|}{P2}                & P1                          & P2                         \\ \hline
GT-Mask $\rightarrow$ Sup-Mask   & 1.091 $\pm$ 0.044   & \multicolumn{1}{c|}{9.358 $\pm$ 0.465}    & 0.916 $\pm$ 0.005              & 7.433 $\pm$ 0.511             \\
GT-Mask $\rightarrow$ Self-Mask  & 1.123 $\pm$ 0.048   & \multicolumn{1}{c|}{9.506 $\pm$ 0.447}    & 0.962 $\pm$ 0.017              & 7.594 $\pm$ 0.442             \\
Sup-Mask $\rightarrow$ Self-Mask & 1.144 $\pm$ 0.038   & \multicolumn{1}{c|}{9.468 $\pm$ 0.286}    & 0.906 $\pm$ 0.026              & 7.365 $\pm$ 0.285             \\ \hlineB{3}
\end{tabular}
\end{adjustbox}
\end{table}

%% file: sections/06_conclusion.tex
\section{Conclusion}
In this paper, by using RPCIN as a probe, we investigated two environment misalignment challenges: \emph{Cross-Domain} and \emph{Cross-Context}. Four datasets: \emph{SimB-Border}, \emph{SimB-Split}, \emph{BlenB-Border}, and \emph{BlenB-Split}, which cover two domains and two contexts, are proposed. Experiment results on the combinations of the proposed datasets reveal potential weaknesses of vison-based long-term dynamics prediction model. Furthermore, to mitigate the \emph{Cross-Domain} challenge, we studied a promising direction and provide an intuitive instance as a concretization, whose effectiveness is demonstrated by empirical results. Lastly, we provide a discussion on the limitations and open issues. 

%% file: sections/08_acknowledgements.tex
\section{Acknowledgements}
\noindent This material is based on research sponsored by Air Force Research Laboratory (AFRL) under agreement number FA8750-19-1-1000. 
The U.S. Government is authorized to reproduce and distribute reprints for Government purposes notwithstanding any copyright notation therein. 
The views and conclusions contained herein are those of the authors and should not be interpreted as necessarily representing the official policies or endorsements, either expressed or implied, of Air Force Laboratory, DARPA or the U.S. Government.

%% file: sections/07_Appendix.tex
\section{Appendix}
\subsection{Dataset Details}
In the paper, we propose four datasets: \emph{SimB-Border}, \emph{SimB-Split}, \emph{BlenB-Border}, and \emph{BlenB-Split} which cover two domains and two contexts. For the datasets on the \emph{Sim} domain, we modified the generation code used in Qi et al.~\yrcite{rpcin} for increasing the image size from 64 $\times$ 64 to 192 $\times$ 96 and introducing borders and split bar as described in \cref{sec:dataset}. 1000 videos with 100 frames in each video are generated for train and test individually. Same with Qi et al.~\yrcite{rpcin}, for each video, only one ball has initial velocities whose magnitude and direction are randomly selected from $\{2, 3, 4, 5, 6\}$ and $\{\frac{i}{6}\pi | i \in [0,1,2,..,11]\}$. For generating dataset in \emph{Blen} domain, by using the ground-truth information of dataset in \emph{Sim} domain as guidance, we represent borders, balls, and split bar as objects in Blender~\cite{blender} and adjust the object properties, such as location, width, and length, and scene properties, such as camera height, for seeking a match between datasets in \emph{Sim} domain and \emph{Blen} domain. We used the \emph{Eevee} as the rendering engine in Blender. 

To find a close match between \emph{Sim} and \emph{Blen} domains, we first create all objects, fix the camera position and direction, rendering engine, and output image resolution. Then, we iteratively adjust the scale and position of objects in Blender and check the output image until finding the unit conversion ratio between Blender (in metric) and output image (in pixel). Finally, we apply the ratio to the ground truth of each \emph{Sim} domain sample for creating the respected sample in the \emph{Blen} domain. However, similar to the real-world scenarios, where data is synthesised to match the real data, since the fundamental mechanisms of image sample generation are different in \emph{Sim} and \emph{Blen} domains, it is arguably infeasible to find an identical match. Furthermore, the slight difference between the two domains can also be considered as one aspect of the domain-specific disturbance inherent in the \emph{Cross-Domain} challenge. By mapping various visual appearances to a common intermediate space, such disturbance can be filtered out, as shown in \cref{tab:sim_to_blen,tab:blen_to_sim}.

\subsection{Training Details}
For the visual backbone, we followed the hourglass~\cite{hourglass} architectures used in RPCIN~\cite{rpcin}, which is a visual feature extractor, that contains a CNN layer and three residual blocks which down-sampled the input image size by a scale of four with output channel size equals to 256, followed by a hourglass module. Details of the model architecture can be found in RPCIN~\cite{rpcin}. For normalization layer, we used BN, IN, GN, and LN, where the group number of GN is set to 32, and LN is implemented by a GN with group number set to one~\cite{groupnorm}. By default, we set the output size of RoI Pooling to three and ablation study results on varying the output size are shown in \cref{fig:ablation}(b). Furthermore, it appears that RPCIN implemented RoI Pooling by using RoI Aligning function of PyTorch which follows \cite{maskrcnn}.

Our experiments are conducted on two NVIDIA 1080ti GPUs. We set total batch size to 40 and models are trained over 50K iterations. We used Adam optimizer~\cite{adam} and set learning rate to $2 \times 10^{-4}$ with cosine learning rate decay~\cite{cosine_decay}. Weight decay is set to $1 \times 10^{-6}$. For input image prepossessing, we only unified the input range to [0, 1] without using dataset specific statistics for further reducing the degree of misalignment between visual domains. Images are randomly flipped horizontally and vertically as augmentation. Same with RPCIN~\cite{rpcin}, each training sample contains four image frames and 24 frames of bounding boxes, where the first four frames of bounding boxes are used for reference and rests are used for prediction ground-truth. Similarly, each testing sample contains four image frames and 44 frames of bounding boxes for evaluating both short-term and long-term prediction. For the best use of dataset, $100-24+1=77$ and $100-44+1=57$ samples can be drawn from a single 100 frames video. Thus, for each dataset, there are 77K samples for train and 57K samples for test.

\subsection{Extracting Segmentation Masks}
For extracting the segmentation masks with either ground-truth label or k-means pseudo label, we adopt a simple encoder-decoder based model. We use the visual feature extractor (prior to the hourglass module) that contains a CNN layer and three residual blocks, as previously described with BN layer, as encoder, which down-samples the size of image by a scale of four with 256 channel outputs. The decoder is a simple five CNN layer with LeakyRelu as activation layer that upsamples the visual feature to the original size with two channels which represent the semantic meaning of a certain location. We use cross-entropy loss for calculating the difference between the prediction and the ground-truth. For self-supervised learning, we use k-means function of OpenCV\cite{opencv_library} to generate pseudo labels for providing supervision signal. Prior to k-means segmentation, for removing the appearance of balls from image that might be mis-classified as border, we first blur the input image and then replace the pixels in those regions that contain balls by the respected pixels in the blurred image. The processed images are used for generating k-means pseudo label. For matching the k-means pseudo label with correct semantic meaning, we assumes that the number of pixels that represent border is less than the number of pixels that represent the table. 

\subsection{Additional Results}
We provide numerical results on model under \emph{Cross-Context} challenge in \cref{tab:board_to_split,tab:split_to_board}, and under both \emph{Cross-Context} and \emph{Cross-Domain} challenges in \cref{tab:cross_both_board,tab:cross_both_split}. For completion, we also conduct experiments on \emph{RealB} dataset~\cite{rpcin} and results are shown in \cref{tab:all_to_real,tab:real_to_all}. However, since the underlying physics mechanism is different between \emph{RealB} and the four proposed dataset, we find it is hard to draw any structural conclusion from the experiments and the results are only provided for reference. 

\input{tables/board_to_split.tex}
\input{tables/split_to_board.tex}

\input{tables/cross_both_board.tex}
\input{tables/cross_both_split.tex}

\input{tables/real_to_all.tex}
\input{tables/all_to_real.tex}

%% file: tables/board_to_split.tex
\begin{table*}[htp]
\centering
\caption{Performance of \emph{Border} context trained RPCIN model on \emph{Cross-Context} challenge with various types of input. Aligned and Cross results on SimB and BlenB  are the same for GT-Mask trained model because they share exactly the same data.} \label{tab:board_to_split}
\begin{adjustbox}{width=0.98\textwidth}
\renewcommand{\arraystretch}{1.3}
\begin{tabular}{ccccccccc}
\hlineB{3}
Source Dataset & \multicolumn{4}{c|}{SimB-Border}                                                                              & \multicolumn{4}{c}{BlenB-Border}                                                         \\ \cline{2-9} 
Target Dataset & \multicolumn{2}{c|}{SimB-Border (\red{Aligned})}                      & \multicolumn{2}{c|}{SimB-Split (\red{Cross})}                       & \multicolumn{2}{c|}{BlenB-Border (\red{Aligned})}                     & \multicolumn{2}{c}{BlenB-Split (\red{Cross})}  \\
Eval Period    & P1             & \multicolumn{1}{c|}{P2}             & P1             & \multicolumn{1}{c|}{P2}              & P1             & \multicolumn{1}{c|}{P2}             & P1             & P2              \\ \hline
\multicolumn{9}{c}{Raw RGB Image Input}                                                                                                                                                                                 \\ \hline
RGB-BN         & 1.131 $\pm$ 0.011 & \multicolumn{1}{c|}{9.568 $\pm$ 0.121} & 6.804 $\pm$ 0.159 & \multicolumn{1}{c|}{39.773 $\pm$ 0.726} & 1.084 $\pm$ 0.023 & \multicolumn{1}{c|}{9.713 $\pm$ 0.554} & 6.587 $\pm$ 0.085 & 40.294 $\pm$ 0.518 \\
RGB-IN         & 1.102 $\pm$ 0.045 & \multicolumn{1}{c|}{9.426 $\pm$ 0.446} & 7.039 $\pm$ 0.164 & \multicolumn{1}{c|}{40.364 $\pm$ 0.483} & 1.103 $\pm$ 0.041 & \multicolumn{1}{c|}{9.471 $\pm$ 0.443} & 6.807 $\pm$ 0.028 & 40.054 $\pm$ 0.600 \\
RGB-GN         & 1.117 $\pm$ 0.058 & \multicolumn{1}{c|}{9.323 $\pm$ 0.346} & 6.721 $\pm$ 0.289 & \multicolumn{1}{c|}{39.356 $\pm$ 0.553} & 1.075 $\pm$ 0.045 & \multicolumn{1}{c|}{9.560 $\pm$ 0.145} & 6.759 $\pm$ 0.087 & 40.223 $\pm$ 0.644 \\
RGB-LN         & 1.085 $\pm$ 0.033 & \multicolumn{1}{c|}{9.165 $\pm$ 0.145} & 6.662 $\pm$ 0.127 & \multicolumn{1}{c|}{39.051 $\pm$ 0.330} & 1.064 $\pm$ 0.020 & \multicolumn{1}{c|}{9.345 $\pm$ 0.350} & 6.762 $\pm$ 0.150 & 39.816 $\pm$ 0.597 \\ \hline
\multicolumn{9}{c}{Segmentation Mask Input}                                                                                                                                                                             \\ \hline
GT-Mask        & 1.091 $\pm$ 0.044 & \multicolumn{1}{c|}{9.358 $\pm$ 0.465} & 8.345 $\pm$ 1.225 & \multicolumn{1}{c|}{39.793 $\pm$ 0.917} & 1.091 $\pm$ 0.044 & \multicolumn{1}{c|}{9.358 $\pm$ 0.465} & 8.345 $\pm$ 1.225 & 39.793 $\pm$ 0.917 \\
Sup-Mask       & 1.093 $\pm$ 0.021 & \multicolumn{1}{c|}{9.396 $\pm$ 0.285} & 8.225 $\pm$ 0.395 & \multicolumn{1}{c|}{40.338 $\pm$ 0.445} & 1.122 $\pm$ 0.036 & \multicolumn{1}{c|}{9.353 $\pm$ 0.268} & 9.271 $\pm$ 1.369 & 40.398 $\pm$ 0.805 \\
Self-Mask     & 1.119 $\pm$ 0.037 & \multicolumn{1}{c|}{9.604 $\pm$ 0.300} & 7.507 $\pm$ 0.237 & \multicolumn{1}{c|}{40.639 $\pm$ 0.908} & 1.136 $\pm$ 0.024 & \multicolumn{1}{c|}{9.945 $\pm$ 0.563} & 8.727 $\pm$ 1.476 & 41.788 $\pm$ 1.302 \\ \hlineB{3}
\end{tabular}
\end{adjustbox}
\end{table*}

%% file: tables/split_to_board.tex
\begin{table*}[htp]
\centering
\caption{Performance of \emph{Split} context trained RPCIN model on \emph{Cross-Context} challenge with various types of input. Aligned and Cross results on SimB and BlenB are the same for GT-Mask trained model because they share exactly the same data.} \label{tab:split_to_board}
\begin{adjustbox}{width=0.98\textwidth}
\renewcommand{\arraystretch}{1.3}
\begin{tabular}{ccccccccc}
\hlineB{3}
Source Dataset & \multicolumn{4}{c|}{SimB-Split}                                                                              & \multicolumn{4}{c}{BlenB-Split}                                                         \\ \cline{2-9} 
Target Dataset & \multicolumn{2}{c|}{SimB-Split (\red{Aligned})}                      & \multicolumn{2}{c|}{SimB-Border (\red{Cross})}                       & \multicolumn{2}{c|}{BlenB-Split (\red{Aligned})}                     & \multicolumn{2}{c}{BlenB-Border (\red{Cross})}  \\
Eval Period    & P1             & \multicolumn{1}{c|}{P2}             & P1             & \multicolumn{1}{c|}{P2}              & P1             & \multicolumn{1}{c|}{P2}             & P1             & P2              \\ \hline
\multicolumn{9}{c}{Raw RGB Image Input}                                                                                                                                                                                 \\ \hline
RGB-BN         & 0.913 $\pm$ 0.019 & \multicolumn{1}{c|}{7.732 $\pm$ 0.208} & 1.589 $\pm$ 0.113 & \multicolumn{1}{c|}{19.238 $\pm$ 0.550} & 0.918 $\pm$ 0.037 & \multicolumn{1}{c|}{7.478 $\pm$ 0.079} & 1.760 $\pm$ 0.265 & 20.492 $\pm$ 0.660 \\
RGB-IN         & 0.945 $\pm$ 0.082 & \multicolumn{1}{c|}{7.641 $\pm$ 0.549} & 3.087 $\pm$ 0.181 & \multicolumn{1}{c|}{28.518 $\pm$ 2.517} & 0.906 $\pm$ 0.062 & \multicolumn{1}{c|}{7.368 $\pm$ 0.168} & 4.083 $\pm$ 0.602 & 30.913 $\pm$ 2.351 \\
RGB-GN         & 0.889 $\pm$ 0.042 & \multicolumn{1}{c|}{7.632 $\pm$ 0.386} & 2.250 $\pm$ 0.067 & \multicolumn{1}{c|}{28.597 $\pm$ 1.720} & 0.931 $\pm$ 0.039 & \multicolumn{1}{c|}{7.641 $\pm$ 0.165} & 2.585 $\pm$ 0.346 & 27.511 $\pm$ 1.150 \\
RGB-LN         & 0.922 $\pm$ 0.042 & \multicolumn{1}{c|}{7.433 $\pm$ 0.237} & 1.683 $\pm$ 0.115 & \multicolumn{1}{c|}{21.756 $\pm$ 1.232} & 0.892 $\pm$ 0.027 & \multicolumn{1}{c|}{7.687 $\pm$ 0.321} & 1.730 $\pm$ 0.156 & 22.934 $\pm$ 1.711 \\ \hline
\multicolumn{9}{c}{Segmentation Mask Input}                                                                                                                                                                             \\ \hline
GT-Mask        & 0.916 $\pm$ 0.005 & \multicolumn{1}{c|}{7.431 $\pm$ 0.511} & 2.085 $\pm$ 0.245 & \multicolumn{1}{c|}{21.713 $\pm$ 2.458} & 0.916 $\pm$ 0.005 & \multicolumn{1}{c|}{7.431 $\pm$ 0.511} & 2.085 $\pm$ 0.245 & 21.713 $\pm$ 2.458 \\
Sup-Mask       & 0.971 $\pm$ 0.011 & \multicolumn{1}{c|}{7.372 $\pm$ 0.089} & 2.006 $\pm$ 0.241 & \multicolumn{1}{c|}{20.220 $\pm$ 1.725} & 0.891 $\pm$ 0.027 & \multicolumn{1}{c|}{7.317 $\pm$ 0.273} & 2.151 $\pm$ 0.401 & 21.798 $\pm$ 2.426 \\
Self-Mask     & 0.911 $\pm$ 0.025 & \multicolumn{1}{c|}{7.837 $\pm$ 1.334} & 2.031 $\pm$ 0.292 & \multicolumn{1}{c|}{21.521 $\pm$ 1.375} & 0.914 $\pm$ 0.022 & \multicolumn{1}{c|}{7.539 $\pm$ 0.227} & 2.433 $\pm$ 0.273 & 21.020 $\pm$ 0.127 \\ \hlineB{3}
\end{tabular}
\end{adjustbox}
\end{table*}

%% file: tables/cross_both_board.tex
\begin{table*}[htp]
\centering
\caption{Performance of \emph{Border} context trained RPCIN model on \emph{Cross-Context} and \emph{Cross-Domain} challenge with various types of input. } \label{tab:cross_both_board}
\begin{adjustbox}{width=0.98\textwidth}
\renewcommand{\arraystretch}{1.3}
\begin{tabular}{ccccccccc}
\hlineB{3}
Source Dataset & \multicolumn{4}{c|}{SimB-Border}                                                                               & \multicolumn{4}{c}{BlenB-Border}                                                              \\ \cline{2-9} 
Target Dataset & \multicolumn{2}{c|}{SimB-Border (\red{Aligned})}                      & \multicolumn{2}{c|}{BlenB-Split (\red{Cross})}                       & \multicolumn{2}{c|}{BlenB-Border (\red{Aligned})}                     & \multicolumn{2}{c}{SimB-Split (\red{Cross})}        \\
Eval Period    & P1             & \multicolumn{1}{c|}{P2}             & P1              & \multicolumn{1}{c|}{P2}              & P1             & \multicolumn{1}{c|}{P2}             & P1                & P2                \\ \hline
\multicolumn{9}{c}{Raw RGB Image Input}                                                                                                                                                                                       \\ \hline
RGB-BN         & 1.131 $\pm$ 0.011 & \multicolumn{1}{c|}{9.568 $\pm$ 0.121} & 12.024 $\pm$ 3.017 & \multicolumn{1}{c|}{48.670 $\pm$ 2.299} & 1.084 $\pm$ 0.023 & \multicolumn{1}{c|}{9.713 $\pm$ 0.554} & 262.567 $\pm$ 75.581 & 228.936 $\pm$ 41.361 \\
RGB-IN         & 1.102 $\pm$ 0.045 & \multicolumn{1}{c|}{9.426 $\pm$ 0.446} & 10.359 $\pm$ 0.664 & \multicolumn{1}{c|}{43.412 $\pm$ 0.736} & 1.103 $\pm$ 0.041 & \multicolumn{1}{c|}{9.471 $\pm$ 0.443} & 28.746 $\pm$ 11.316  & 67.175 $\pm$ 13.291  \\
RGB-GN         & 1.117 $\pm$ 0.058 & \multicolumn{1}{c|}{9.323 $\pm$ 0.346} & 7.072 $\pm$ 0.267  & \multicolumn{1}{c|}{40.172 $\pm$ 0.731} & 1.075 $\pm$ 0.045 & \multicolumn{1}{c|}{9.560 $\pm$ 0.145} & 8.217 $\pm$ 0.242    & 42.131 $\pm$ 1.377   \\
RGB-LN         & 1.085 $\pm$ 0.033 & \multicolumn{1}{c|}{9.165 $\pm$ 0.145} & 7.729 $\pm$ 0.671  & \multicolumn{1}{c|}{41.274 $\pm$ 1.092} & 1.064 $\pm$ 0.020 & \multicolumn{1}{c|}{9.345 $\pm$ 0.350} & 14.404 $\pm$ 1.762   & 53.815 $\pm$ 2.412   \\ \hline
\multicolumn{9}{c}{Segmentation Mask Input}                                                                                                                                                                                   \\ \hline
GT-Mask        & 1.091 $\pm$ 0.044 & \multicolumn{1}{c|}{9.358 $\pm$ 0.465} & 8.345 $\pm$ 1.225  & \multicolumn{1}{c|}{39.793 $\pm$ 0.917} & 1.091 $\pm$ 0.044 & \multicolumn{1}{c|}{9.358 $\pm$ 0.465} & 8.345 $\pm$ 1.225    & 39.793 $\pm$ 0.917   \\
Sup-Mask       & 1.093 $\pm$ 0.021 & \multicolumn{1}{c|}{9.396 $\pm$ 0.285} & 8.259 $\pm$ 0.394  & \multicolumn{1}{c|}{40.414 $\pm$ 0.420} & 1.122 $\pm$ 0.036 & \multicolumn{1}{c|}{9.353 $\pm$ 0.268} & 9.232 $\pm$ 1.356    & 40.303 $\pm$ 0.748   \\
Self-Mask     & 1.119 $\pm$ 0.037 & \multicolumn{1}{c|}{9.604 $\pm$ 0.300} & 7.712 $\pm$ 0.282  & \multicolumn{1}{c|}{40.893 $\pm$ 0.951} & 1.136 $\pm$ 0.024 & \multicolumn{1}{c|}{9.945 $\pm$ 0.563} & 8.252 $\pm$ 1.111    & 41.206 $\pm$ 1.113   \\ \hlineB{3}
\end{tabular}
\end{adjustbox}
\end{table*}

%% file: tables/cross_both_split.tex
\begin{table*}[htp]
\centering
\caption{Performance of \emph{Split} context trained RPCIN model on \emph{Cross-Context} and \emph{Cross-Domain} challenge with various types of input. } \label{tab:cross_both_split}
\begin{adjustbox}{width=0.98\textwidth}
\renewcommand{\arraystretch}{1.3}
\begin{tabular}{ccccccccc}
\hlineB{3}
Source Dataset & \multicolumn{4}{c|}{SimB-Split}                                                                                & \multicolumn{4}{c}{BlenB-Split}                                                              \\ \cline{2-9} 
Target Dataset & \multicolumn{2}{c|}{SimB-Split (\red{Aligned})}                      & \multicolumn{2}{c|}{BlenB-Border (\red{Cross})}                        & \multicolumn{2}{c|}{BlenB-Split (\red{Cross})}                     & \multicolumn{2}{c}{SimB-Border (\red{Cross})}        \\
Eval Period    & P1             & \multicolumn{1}{c|}{P2}             & P1              & \multicolumn{1}{c|}{P2}               & P1             & \multicolumn{1}{c|}{P2}             & P1                & P2                \\ \hline
\multicolumn{9}{c}{Raw RGB Image Input}                                                                                                                                                                                        \\ \hline
RGB-BN         & 0.913 $\pm$ 0.019 & \multicolumn{1}{c|}{7.732 $\pm$ 0.208} & 9.471 $\pm$ 3.328  & \multicolumn{1}{c|}{43.641 $\pm$ 8.771}  & 0.918 $\pm$ 0.037 & \multicolumn{1}{c|}{7.478 $\pm$ 0.079} & 171.341 $\pm$ 79.239 & 193.574 $\pm$ 45.798 \\
RGB-IN         & 0.945 $\pm$ 0.082 & \multicolumn{1}{c|}{7.641 $\pm$ 0.549} & 12.034 $\pm$ 2.199 & \multicolumn{1}{c|}{52.677 $\pm$ 5.350}  & 0.906 $\pm$ 0.062 & \multicolumn{1}{c|}{7.368 $\pm$ 0.168} & 29.570 $\pm$ 14.694  & 70.004 $\pm$ 15.201  \\
RGB-GN         & 0.889 $\pm$ 0.042 & \multicolumn{1}{c|}{7.632 $\pm$ 0.386} & 8.250 $\pm$ 2.052  & \multicolumn{1}{c|}{47.316 $\pm$ 3.474}  & 0.931 $\pm$ 0.039 & \multicolumn{1}{c|}{7.641 $\pm$ 0.165} & 8.721 $\pm$ 1.168    & 44.849 $\pm$ 3.762   \\
RGB-LN         & 0.922 $\pm$ 0.042 & \multicolumn{1}{c|}{7.433 $\pm$ 0.237} & 8.251 $\pm$ 3.202  & \multicolumn{1}{c|}{46.966 $\pm$ 10.581} & 0.892 $\pm$ 0.027 & \multicolumn{1}{c|}{7.687 $\pm$ 0.321} & 12.106 $\pm$ 1.250   & 57.674 $\pm$ 2.721   \\ \hline
\multicolumn{9}{c}{Segmentation Mask Input}                                                                                                                                                                                    \\ \hline
GT-Mask        & 0.916 $\pm$ 0.005 & \multicolumn{1}{c|}{7.431 $\pm$ 0.511} & 2.085 $\pm$ 0.245  & \multicolumn{1}{c|}{21.713 $\pm$ 2.458}  & 0.916 $\pm$ 0.005 & \multicolumn{1}{c|}{7.431 $\pm$ 0.511} & 2.085 $\pm$ 0.245    & 21.713 $\pm$ 2.458   \\
Sup-Mask       & 0.971 $\pm$ 0.011 & \multicolumn{1}{c|}{7.372 $\pm$ 0.089} & 2.007 $\pm$ 0.240  & \multicolumn{1}{c|}{20.224 $\pm$ 1.723}  & 0.891 $\pm$ 0.027 & \multicolumn{1}{c|}{7.317 $\pm$ 0.273} & 2.151 $\pm$ 0.401    & 21.799 $\pm$ 2.426   \\
Self-Mask     & 0.911 $\pm$ 0.025 & \multicolumn{1}{c|}{7.837 $\pm$ 1.334} & 2.053 $\pm$ 0.291  & \multicolumn{1}{c|}{21.519 $\pm$ 1.390}  & 0.914 $\pm$ 0.022 & \multicolumn{1}{c|}{7.539 $\pm$ 0.227} & 2.428 $\pm$ 0.274    & 21.021 $\pm$ 0.113   \\ \hlineB{3}
\end{tabular}
\end{adjustbox}
\end{table*}

%% file: tables/real_to_all.tex
\begin{table*}[htp]
\centering
\caption{Performance of realB trained RPCIN model on \emph{Cross-Context} and \emph{Cross-Domain} challenge with various types of input. } \label{tab:real_to_all}
\begin{adjustbox}{width=0.98\textwidth}
\renewcommand{\arraystretch}{1.3}
\begin{tabular}{ccccccccccl}
\hlineB{3}
Source Dataset & \multicolumn{10}{c}{RealB}                                                                                                                                                                                                                                                \\ \cline{2-11} 
Target Dataset & \multicolumn{2}{c|}{RealB (\red{Aligned}) }                           & \multicolumn{2}{c|}{SimB-Border (\red{Cross}) }                         & \multicolumn{2}{c|}{SimB-Split (\red{Cross})}                         & \multicolumn{2}{c|}{BlenB-Border (\red{Cross}) }                         & \multicolumn{2}{c}{BlenB-Split (\red{Cross}) }     \\
Eval Period    & P1             & \multicolumn{1}{c|}{P2}             & P1              & \multicolumn{1}{c|}{P2}               & P1               & \multicolumn{1}{c|}{P2}              & P1               & \multicolumn{1}{c|}{P2}               & P1               & P2               \\ \hline
\multicolumn{11}{c}{Raw RGB Image Input}                                                                                                                                                                                                                                                   \\ \hline
RGB-BN         & 0.421 $\pm$ 0.009 & \multicolumn{1}{c|}{3.004 $\pm$ 0.091} & 28.604 $\pm$ 3.428 & \multicolumn{1}{c|}{67.030 $\pm$ 4.847}  & 37.838 $\pm$ 10.681 & \multicolumn{1}{c|}{81.673 $\pm$ 9.417} & 49.760 $\pm$ 19.008 & \multicolumn{1}{c|}{84.241 $\pm$ 19.333} & 52.235 $\pm$ 17.935 & 91.045 $\pm$ 16.429 \\
RGB-IN         & 0.549 $\pm$ 0.020 & \multicolumn{1}{c|}{3.598 $\pm$ 0.173} & 12.492 $\pm$ 1.948 & \multicolumn{1}{c|}{48.657 $\pm$ 6.188}  & 15.412 $\pm$ 1.563  & \multicolumn{1}{c|}{56.800 $\pm$ 4.458} & 9.121 $\pm$ 0.370   & \multicolumn{1}{c|}{41.226 $\pm$ 0.710}  & 15.324 $\pm$ 0.466  & 55.429 $\pm$ 1.356  \\
RGB-GN         & 0.423 $\pm$ 0.017 & \multicolumn{1}{c|}{3.050 $\pm$ 0.304} & 9.055 $\pm$ 0.993  & \multicolumn{1}{c|}{74.546 $\pm$ 58.718} & 12.413 $\pm$ 0.970  & \multicolumn{1}{c|}{54.200 $\pm$ 3.539} & 8.091 $\pm$ 0.376   & \multicolumn{1}{c|}{39.766 $\pm$ 1.036}  & 11.413 $\pm$ 0.604  & 51.350 $\pm$ 1.351  \\
RGB-LN         & 0.416 $\pm$ 0.038 & \multicolumn{1}{c|}{3.163 $\pm$ 0.268} & 8.669 $\pm$ 2.168  & \multicolumn{1}{c|}{40.725 $\pm$ 3.583}  & 12.341 $\pm$ 1.688  & \multicolumn{1}{c|}{54.137 $\pm$ 1.208} & 8.366 $\pm$ 1.412   & \multicolumn{1}{c|}{41.000 $\pm$ 2.947}  & 12.527 $\pm$ 1.596  & 57.103 $\pm$ 2.933  \\ \hlineB{3}
\end{tabular}
\end{adjustbox}
\end{table*}

%% file: tables/all_to_real.tex
\begin{table*}[htp]
\centering
\caption{Performance of various datasets trained RPCIN models on challenge with input as RealB. } \label{tab:all_to_real}
\begin{adjustbox}{width=0.98\textwidth}
\renewcommand{\arraystretch}{1.3}
\begin{tabular}{ccccccccc}
\hlineB{3}
Source Dataset & \multicolumn{2}{c|}{SimB-Border}                        & \multicolumn{2}{c|}{SimB-Split}                         & \multicolumn{2}{c|}{BlenB-Border}                        & \multicolumn{2}{c}{BlenB-Split}   \\ \cline{2-9} 
Target Dataset & \multicolumn{8}{c}{RealB (\red{Cross}) }                                                                                                                                                                                      \\ \hline
Eval Period    & P1              & \multicolumn{1}{c|}{P2}              & P1              & \multicolumn{1}{c|}{P2}               & P1              & \multicolumn{1}{c|}{P2}               & P1              & P2              \\ \hline
\multicolumn{9}{c}{Raw RGB Image Input}                                                                                                                                                                                         \\ \hline
RGB-BN         & 23.697 $\pm$ 1.548 & \multicolumn{1}{c|}{39.668 $\pm$ 6.773} & 14.651 $\pm$ 3.376 & \multicolumn{1}{c|}{32.408 $\pm$ 4.780}  & 32.291 $\pm$ 2.968 & \multicolumn{1}{c|}{65.397 $\pm$ 16.150} & 28.887 $\pm$ 4.330 & 47.890 $\pm$ 9.987 \\
RGB-IN         & 7.746 $\pm$ 0.981  & \multicolumn{1}{c|}{29.736 $\pm$ 4.457} & 12.424 $\pm$ 3.594 & \multicolumn{1}{c|}{42.159 $\pm$ 10.807} & 3.181 $\pm$ 0.885  & \multicolumn{1}{c|}{16.514 $\pm$ 2.280}  & 8.243 $\pm$ 4.815  & 32.189 $\pm$ 9.774 \\
RGB-GN         & 2.510 $\pm$ 0.232  & \multicolumn{1}{c|}{11.921 $\pm$ 1.321} & 4.791 $\pm$ 0.879  & \multicolumn{1}{c|}{21.269 $\pm$ 2.668}  & 2.166 $\pm$ 0.415  & \multicolumn{1}{c|}{12.578 $\pm$ 0.332}  & 4.683 $\pm$ 0.761  & 21.258 $\pm$ 2.482 \\
RGB-LN         & 3.208 $\pm$ 0.947  & \multicolumn{1}{c|}{14.067 $\pm$ 3.712} & 4.447 $\pm$ 0.567  & \multicolumn{1}{c|}{23.891 $\pm$ 5.079}  & 4.661 $\pm$ 1.928  & \multicolumn{1}{c|}{19.314 $\pm$ 7.201}  & 2.447 $\pm$ 0.478  & 14.576 $\pm$ 2.699 \\ \hlineB{3}
\end{tabular}
\end{adjustbox}
\end{table*}